\documentclass[runningheads]{llncs}

\usepackage[T1]{fontenc}

\usepackage{graphicx}
\usepackage{comment}
\usepackage{amsmath,amssymb}
\usepackage{color}
\usepackage{url}
\usepackage{hyperref}
\usepackage{subcaption}
\usepackage{multirow}

\newif\ifreview
\reviewfalse

\ifreview
	\usepackage{lineno}

	\linenumbers
\fi

\begin{document}


\def\SubNumber{021}

\def\GCPRTrack{Main Track}

\title{SegSLR: Promptable Video Segmentation for Isolated Sign Language Recognition}

\ifreview
\else
	\titlerunning{SegSLR for Isolated Sign Language Recognition}

	\author{Sven Schreiber\orcidID{0009-0009-8652-9549} \and
	Noha Sarhan\orcidID{0000-0002-1545-9346} \and
	Simone Frintrop\orcidID{0000-0002-9475-3593} \and
    Christian Wilms\orcidID{0009-0003-2490-7029}}
	
	
	\institute{Computer Vision Group, University of Hamburg, Germany
	\email{firstname.lastname@uni-hamburg.de}\\
	}
\fi

\maketitle              

\begin{abstract}
Isolated Sign Language Recognition (ISLR) approaches primarily rely on RGB data or signer pose information. However, combining these modalities often results in the loss of crucial details, such as hand shape and orientation, due to imprecise representations like bounding boxes. 
Therefore, we propose the ISLR system SegSLR, which combines RGB and pose information through promptable zero-shot video segmentation. Given the rough localization of the hands and the signer's body from pose information, we segment the respective parts through the video to maintain all relevant shape information. Subsequently, the segmentations focus the processing of the RGB data on the most relevant body parts for ISLR. This effectively combines RGB and pose information. Our evaluation on the complex ChaLearn249 IsoGD dataset shows that SegSLR outperforms state-of-the-art methods. Furthermore, ablation studies indicate that SegSLR strongly benefits from focusing on the signer's body and hands, justifying our design choices.  

\end{abstract}

\section{Introduction}
\label{sec:intro}

Sign language is a central way to communicate for the deaf or hard-of-hearing. It transmits information through several visual parameters, most importantly manual parameters like hand shape, orientation, position, or movement, but also non-manual parameters like body posture, facial expressions, or head movements~\cite{baker1991american}. Outside the deaf or hard-of-hearing community, few people understand sign language, which substantially limits the social interaction of the deaf or hard-of-hearing. To bridge this gap, Isolated Sign Language Recognition~(ISLR) systems classify a video sequence of sign language on a gloss-level. 

ISLR systems rely on various techniques~\cite{rastgoo2021sign,sarhan2023unraveling}. Several systems explicitly focus on the most relevant image areas like the signer's body, hands, or face to capture the manual and non-manual parameters~\cite{gokcce2020score,hosain2021hand,decoster2021isolated,sarhan2023hands,sarhan2023pseudodepth,hu2021signbert,lee2023human}. This is done by attending to the relevant parts of the RGB video frames~(RGB-based) or adding dedicated networks, which encode pose information~(pose-based). Pose-based models extract keypoints of the signer and subsequently generate a skeleton-like graph, which is then processed by complex architectures~\cite{lin2024skim,hu2021signbert,lee2023human,tunga2021pose}.
In contrast, RGB-based models mostly extract crops around the signer's hands or the signer itself. These crops are classified as one part of the system. Note that some RGB-based methods use pose information to locate the relevant parts in the image. Thus, most methods either operate solely on RGB or pose information, or use the pose mostly to guide the extraction of crops, losing critical details about the pose, such as hand shape, hand orientation, or body posture.

Recently, foundation models for promptable segmentation of images and videos revolutionized segmentation in several domains and tasks~\cite{cheng2023tracking,wu2023medical,ma2024segment,he2024weakly,wilms2024sos}. Promptable segmentation refers to segmenting a coherent part of the image given a suitable prompt, like point coordinates or a bounding box. SAM~\cite{kirillov2023segment} for images and SAM~2~\cite{ravi2024sam} for videos, address these tasks and generate high-quality segmentations in a zero-shot manner without domain-specific training data. Hence, these models are also well-suited for ISLR.


\begin{figure}[tb]
    \centering
    \includegraphics[width=0.7\textwidth]{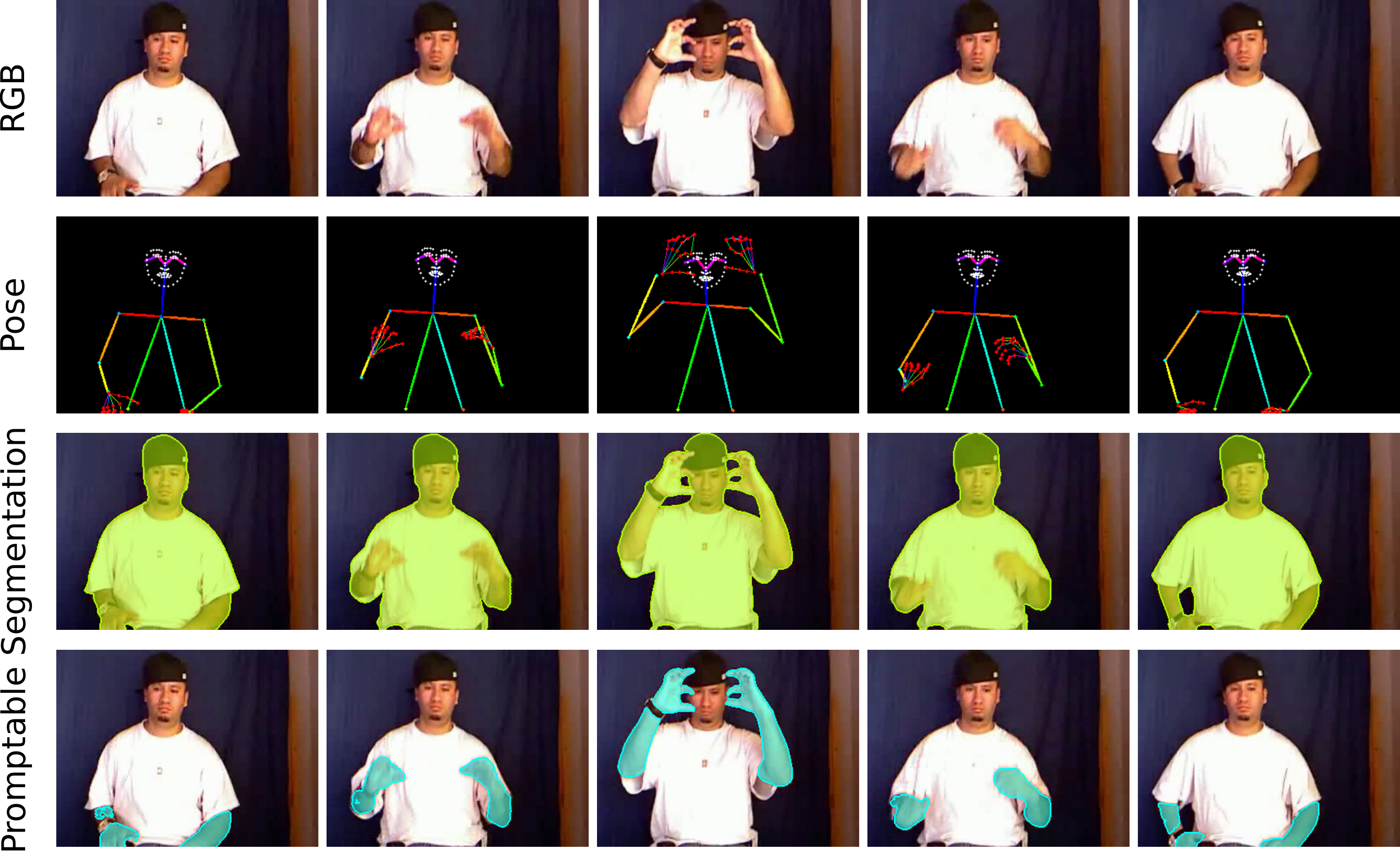}
    \caption{Idea of the proposed SegSLR system: We combine RGB information~(first row) and pose information~(second row) through promptable video segmentation. The pose information is used as prompts to segment the RGB frames. This leads to segmentations of the signer's body~(third row) and hands~(fourth row) to focus processing on the most relevant image areas for ISLR.}
    \label{fig:cover}
\end{figure}

In this paper, we combine RGB-based and pose-based ISLR through the innovative use of promptable video segmentation in our novel ISLR system SegSLR, visualized in Fig.~\ref{fig:system}. SegSLR uses a multi-stream approach. Besides Inflated 3D CNN (I3D CNN)~\cite{carreira2017quo} based streams for plain RGB information and optical flow, ensuring a comprehensive understanding of both spatial and temporal dynamics, SegSLR contains segmentation streams, which combine RGB and pose information through promptable video segmentation. The innovative idea of these streams is, first, to utilize pose information to locate the signer's body and hands in the form of point sets. Second, these point sets are utilized by the promptable video segmentation method SAM~2 to generate high-quality segmentations~(masklets) of the signer's body and hands through the video in a zero-shot manner to retain rich pose information~(see Fig.~\ref{fig:cover} for an example). Finally, these masklets are used to focus the processing of the RGB frames in the subsequent classifiers based on I3D CNNs to the most important image areas for ISLR. This results in a unified architecture exclusively relying on simple I3D CNNs for classification. Based on this innovative combination of pose and RGB information, SegSLR outperforms state-of-the-art systems on the complex ChaLearn249 IsoGD dataset, as our evaluation shows. Ablation studies also demonstrate the benefit of the key design decisions.

Overall, our contributions are three-fold:
\begin{itemize}
    \item We introduce SegSLR, a novel ISLR method,  leveraging both RGB and pose information.
    \item We use foundation models for promptable video segmentation in ISLR to combine pose and RGB information, which retains rich pose information through high-quality segmentations.
    \item SegSLR outperforms the existing state-of-the-art by up to 4.17\%, while ablation studies validate the design choices.
\end{itemize}

\section{Related Work}

\label{sec:relWork_ISLR}
Methods for ISLR evolved from models based on hand-crafted features to learned CNN-, LSTM-, or transformer-based architectures~\cite{rastgoo2021sign,sarhan2023unraveling}. Here, we will discuss learned methods since they substantially outperform traditional ones. Learned models for ISLR can be divided into three groups: methods that mainly rely on RGB information, on pose information, or hybrid methods.

Methods relying only on RGB input either encode the entire frame or, additionally, crops of relevant areas. \cite{sarhan2020transfer}~encode entire RGB frames of a video in an I3D CNN architecture~\cite{carreira2017quo} and add a second stream with optical flow to better capture the temporal dynamics. Subsequently, the streams are fused on score-level. Similarly, \cite{li2020transferring}~apply I3D CNN to the RGB input and add weakly supervised data to improve the feature extraction. \cite{zhu2017multimodal}~combine 3D CNNs for short-term temporal dependencies with a Conv LSTM for long-term ones.  


Several RGB-based methods additionally focus on the signer's body parts as important parameters for sign language recognition. \cite{lim2019isolated}~apply simple hand detectors based on Haar-like features to extract bounding boxes around the hands and track them through the video, yielding a hand energy image for classification. \cite{sarhan2023hands,sarhan2021spatial,sarhan2023pseudodepth}~all extend the two-stream model of \cite{sarhan2020transfer} by focusing on different aspects. While \cite{sarhan2023hands} extract masks of the hands in an individual stream to mask the RGB input, \cite{sarhan2021spatial} focus the RGB input on the moving areas in an end-to-end learned manner, resulting in attending hands and arms. Finally, \cite{sarhan2023pseudodepth} utilize pseudo depth as an additional stream, effectively segmenting the signer without combining this information with the RGB stream.


Using only pose information, \cite{lin2024skim} apply a graph convolutional network~(GCN) to the skeleton graph generated by a pose estimation system, while also applying advanced data augmentations, which mix signs. Also utilizing GCNs as their backbone, \cite{wong2023learnt} add text embeddings in a contrastive learning framework, while \cite{tunga2021pose} use GCNs at the frame-level and model the temporal dependency using BERT~\cite{devlin2018bert}. \cite{hu2021signbert,zhao2023best} use BERT for self-supervised pre-training. In \cite{hu2021signbert}, masked hand poses are reconstructed for pre-training. In contrast to previous methods, \cite{lee2023human} apply transformer-based modules instead of GCNs. 


Hybrid methods combine RGB and pose information. \cite{jiang2021skeleton}~propose a multi-stream architecture with 3D CNNs for RGB and optical flow input as well as a GCN for the skeleton graph, combining them through score fusion. Several systems \cite{gokcce2020score,hosain2021hand,gruber2021mutual,decoster2021isolated} crop parts of the RGB frames or latent representations around the hands, face, or the entire signer for focused processing based on pose information. Yet, only bounding boxes \cite{gokcce2020score,hosain2021hand,decoster2021isolated} or rough segmentations~\cite{gruber2021mutual} are used. Finally, \cite{zuo2023natural} encode the position of specific joints of the signer into heatmaps that can be processed by 3D CNNs, which also capture RGB information~\cite{jiang2021skeleton}.




Closest to our proposed SegSLR are the hybrid methods, which use pose information to focus processing of the RGB data on relevant areas like the signer's body or hands. However, \cite{gokcce2020score,hosain2021hand,decoster2021isolated}~only use bounding boxes, losing information about the hand shape, hand orientation, or the signer's body posture, which are all highly relevant for sign language recognition. \cite{gruber2021mutual}~generate pixel-precise segmentations, yet they are derived from the pose information through dilation. This results in low-quality segmentation masks. In contrast, SegSLR employs high-quality segmentations of the signer's body and hands using a foundation model, retaining important details about body posture and manual parameters.

\section{Revisit Segment Anything Model 2}
Original SAM~\cite{kirillov2023segment} for images is a segmentation system trained on a large-scale dataset, which is applied to various tasks in a zero-shot manner. To guide SAM, prompts~(points, boxes, or masks) are used, which indicate what to segment within an image. SAM~2~\cite{ravi2024sam} extends this idea to videos and leverages the temporal information to segment consistent masklets. SAM~2 consists of three major components: encoders, a decoder, and a memory mechanism. The image and prompt encoders consist of a masked autoencoder~\cite{he2022maskedae} for images as well as positional and learned embeddings for the prompts. Given these embeddings, the mask decoder predicts a segmentation mask and an IoU score. To ensure temporal consistency across the video, a memory attention mechanism is added.  For training SAM~2, \cite{ravi2024sam}~proposed a new dataset SA-V, with manually or semi-automatically annotated videos.

\section{Method}
\begin{figure*}[tb]
    \centering
    \includegraphics[width=\textwidth]{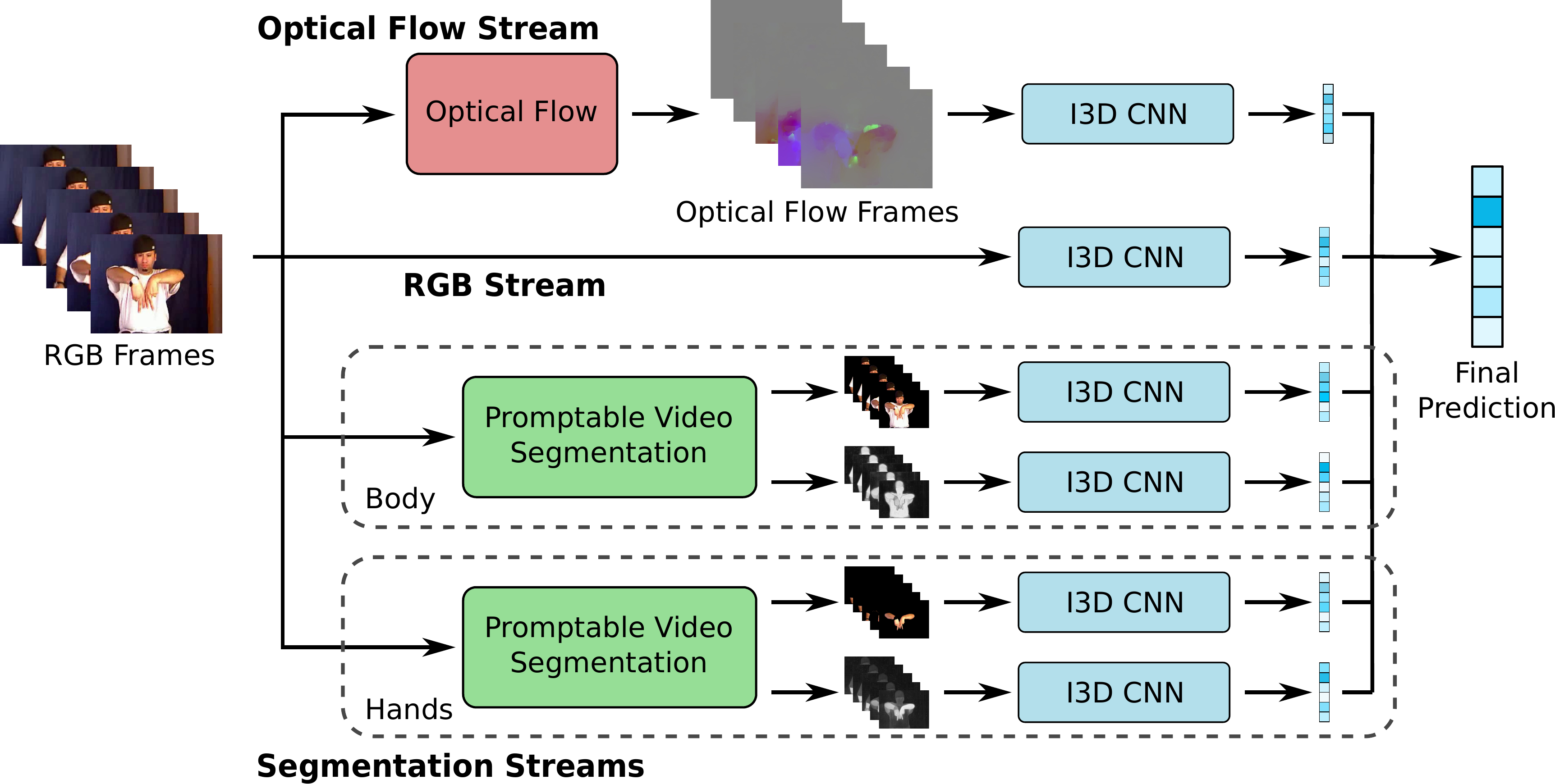}
    \caption{Overview of our proposed SegSLR system. Based on RGB frames of a video, the first stream~(Optical Flow Stream) calculates the optical flow and, subsequently, classifies these optical flow frames using an I3D CNN. The second stream, RGB Stream, directly applies an I3D CNN to the plain RGB frames. As the key novelty of SegSLR, we propose the segmentation streams, which combine RGB and pose information using promptable video segmentation modules~(see Fig.~\ref{fig:video_seg}) to generate four outputs: RGB frames focused on the signer's body and hands as well as the respective segmentation logits. These outputs are processed by I3D CNNs. Finally, score-level fusion is applied to aggregate the results. }
    \label{fig:system}
\end{figure*}

This section introduces our novel SegSLR system for ISLR, as depicted in Fig.~\ref{fig:system}. SegSLR follows the general approach of~\cite{sarhan2020transfer}, comprising a multi-stream architecture based on Inflated 3D CNNs~(I3D CNNs)~\cite{carreira2017quo} with streams for processing plain RGB frames and derived optical flow information, as visible in the center and top of Fig.~\ref{fig:system}. In addition, SegSLR has four segmentation streams (see bottom part in Fig.~\ref{fig:system}) to effectively combine RGB and pose information. To this end, a promptable video segmentation module as visualized in Fig.~\ref{fig:video_seg} first estimates the signer's pose and generates keypoints for the signer's body and hands. A selection of these keypoints is used to prompt SAM~2, yielding high-quality segmentation masks of the signer's body and hands. Subsequently, two streams in SegSLR directly take the logits of the segmentation masks for the body and both hands, respectively. The other two segmentation streams mask the RGB frames, to focus processing on the relevant parts of the RGB input while maintaining rich pose information~(hand shape, hand orientation, and body posture) through the high-quality segmentations. Finally, the data of each stream is processed by an I3D CNN per stream, and the results of all I3D CNNs are combined using score-level fusion~(see right part in Fig.~\ref{fig:system}). In the following sections, we discuss each step in more detail.

\subsection{Pose Estimation and Prompt Generation}
\label{sec:prompt}

\begin{figure*}[tb]
    \centering
    \begin{subfigure}[b]{0.48\textwidth}
        \includegraphics[width=\textwidth]{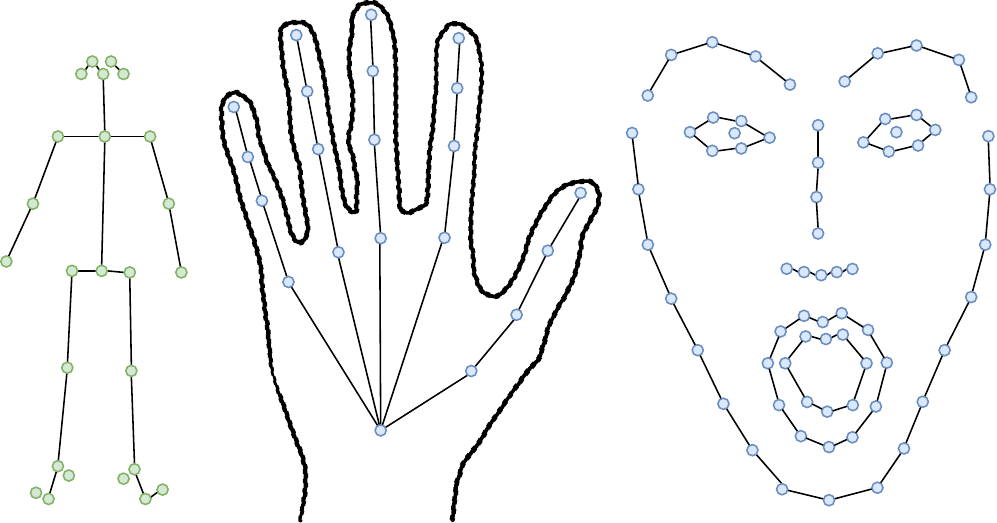}
        \caption{Selected keypoints for the body}
        \label{fig:keypoints_body}
    \end{subfigure}
    ~
    \begin{subfigure}[b]{0.48\textwidth}
        \includegraphics[width=\textwidth]{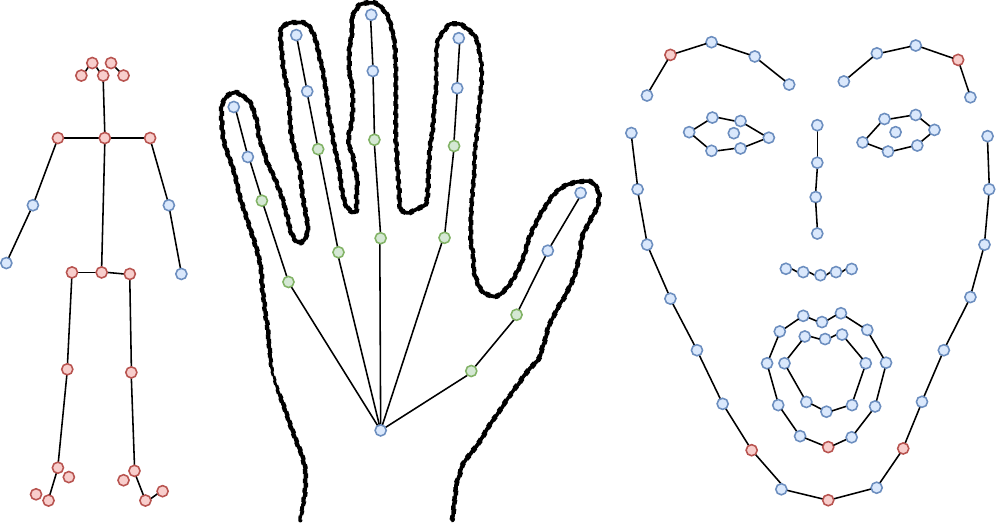}
        \caption{Selected keypoints for the hands}
        \label{fig:keypoints_hand}
    \end{subfigure}
    \caption{Overview of RTMW~\cite{jiang2024rtmw} keypoints for prompting SAM~2 to segment the signer's body~(\subref{fig:keypoints_body}) and hands~(\subref{fig:keypoints_hand}). Green keypoints are positive point prompts, red keypoints are negative point prompts, and blue keypoints are ignored. 
    }
    \label{fig:keypoints}
\end{figure*}

\begin{figure}[tb]
    \centering
    \includegraphics[width=0.7\textwidth]{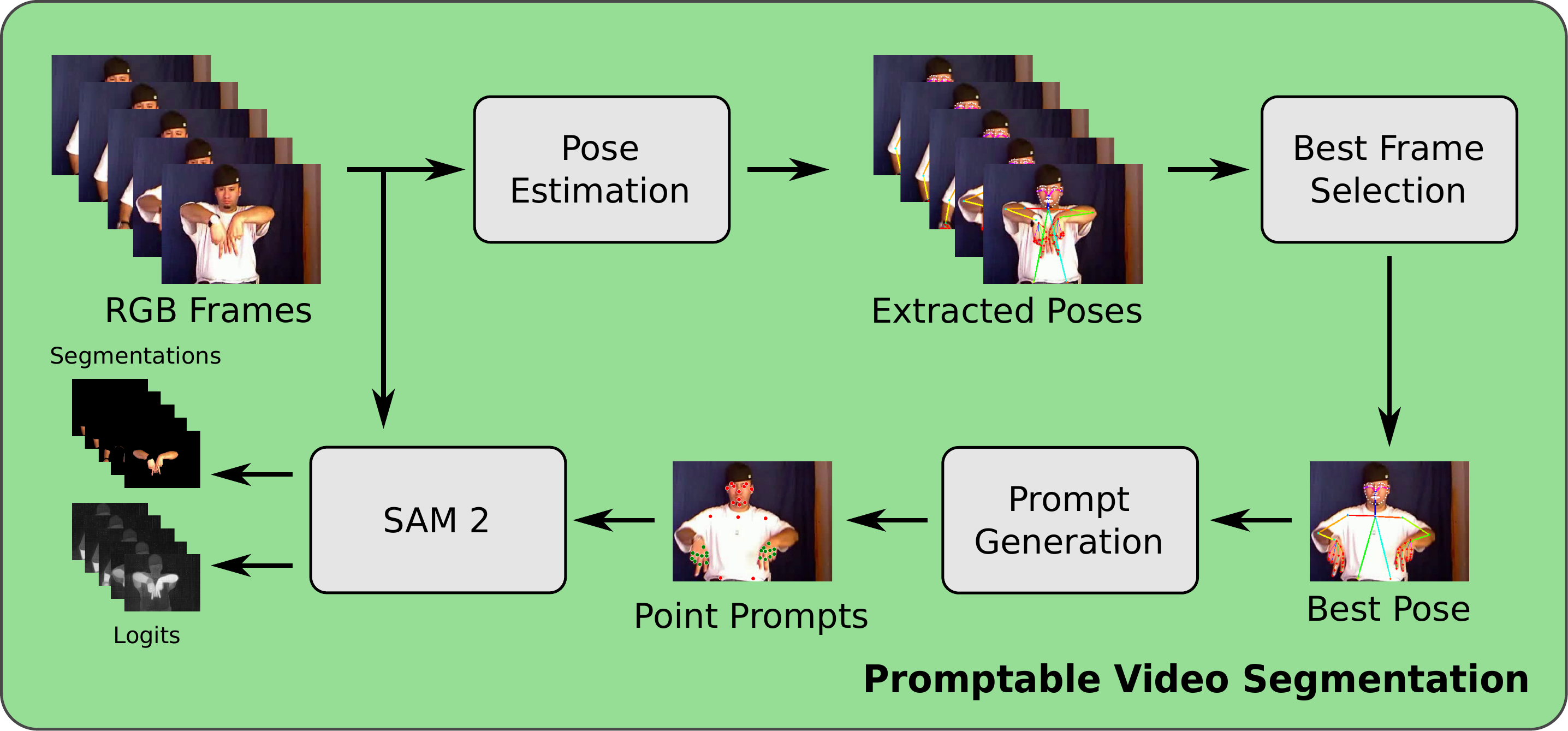}
    \caption{Detailed view of our promptable video segmentation module for hands segmentation. For segmenting the signer's body, the same pipeline is applied with different point prompts. From the input RGB frames, the signer's pose is estimated, and the best frame to initiate the video segmentation is determined. In this frame, several keypoints are converted to positive or negative point prompts. Using these prompts, SAM~2 is applied to the RGB frames. The results are the RGB frames masked by the SAM~2 segmentations and the segmentations' logits. Note that the estimated poses and the point prompts are overlayed with the RGB frames for visualisation only.}
    \label{fig:video_seg}
\end{figure}

As a first step in all segmentation streams of SegSLR, we derive point prompts based on pose information for prompting SAM~2 in our promptable video segmentation module~(see Fig.~\ref{fig:video_seg}). As outlined in Sec.~\ref{sec:intro}, pose estimation results are used in ISLR to capture the body posture or locate the hands~\cite{gokcce2020score,hosain2021hand,gruber2021mutual,decoster2021isolated}. Therefore, we take the state-of-the-art human pose estimation system RTMW~\cite{jiang2024rtmw} and initially extract 116 keypoints, as visualized in Fig.~\ref{fig:keypoints}, covering the entire signer's body. Given these keypoints, we create two subsets. The first subset~(see Fig.~\ref{fig:keypoints_body}) captures the entire signer's body, therefore, we use all keypoints except for detailed hand and face keypoints, since this level of detail is not necessary. The second subset focuses on the hands and includes the keypoints around the first joint per finger, as visualized in Fig.~\ref{fig:keypoints_hand}. We ignore the other hand keypoints since they are frequently outside the hands due to imprecise localization. We also select negative keypoints to discern the hands from the remaining body in SAM-2. As negative keypoints, we select all major body keypoints and important face keypoints. All keypoints in both subsets are the point prompts to guide SAM~2. Note that undetected keypoints are ignored for prompt generation.

\subsection{Best Frame Selection}

Before applying SAM~2, we select the best frame to start the segmentation through the video in our promptable video segmentation module. Frame selection is important since the first frame might not cover the hands, as the signer's hands are usually at hip level. For instance, this is apparent from the initial frame in Fig.~\ref{fig:cover}. To select the best frame, we assess the quality of the detected keypoints per frame by calculating the average keypoint confidence, the size of the bounding box around all keypoints, and the overlap between hands and face keypoints. The keypoint confidence is a per-keypoint output from RTMW and a surrogate for RTMW's confidence about the keypoint. To calculate a single score, we take the average across all detected keypoints. The remaining two measures are used to select a frame, which shows no or only a minimal overlap between the hands and the face. Hands in front of the face frequently occur in sign language. However, due to their similarity in color, it is challenging to discern these regions for a segmentation system when they overlap. Therefore, we first calculate the area of the bounding box covering all keypoints. This helps to measure how sprawled the arms are. To prevent cases where only the upper arms are stretched out and the hands are still close to the face, we determine the maximum overlap between one hand's bounding box and the bounding box around the face. Note that all bounding boxes are created based on the detected RTMW keypoints. To combine the three scores, we first normalize the scores by their respective per-video maximum, subtract the maximum overlap from 1, and multiply them. Hence, the best frame will receive the maximum score. Given this frame, we use the respective sets of point prompts for prompting SAM~2.


\subsection{Mask Generation}

Given the two sets of point prompts for the signer's body and hands on the best frame of a video sequence, we bidirectionally prompt SAM~2 starting from the best frame in our promptable video segmentation module. Hence, we apply SAM~2 from the best frame forward through the video and backward. This process is conducted for each set. The resulting masklet per set is applied to the RGB frames to focus on either the signer's body or hands. Besides the binary segmentation masks, we also extract the per-frame logits of the segmentations to capture the global scene context through SAM~2's per-pixel confidence.


\subsection{ISLR Classification}

\begin{figure*}[tb]
    \centering
    \begin{subfigure}[b]{0.23\textwidth}
        \includegraphics[width=\textwidth]{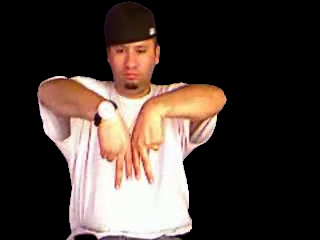}
        \caption{Body segmentation}
        \label{fig:seg_example_body_mask}
    \end{subfigure}
    ~
    \begin{subfigure}[b]{0.23\textwidth}
        \includegraphics[width=\textwidth]{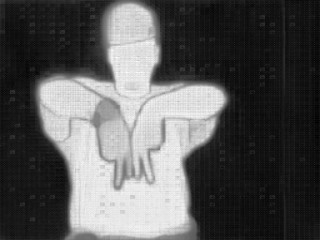}
        \caption{Body logits\\ \ }
        \label{fig:seg_example_body_logits}
    \end{subfigure}
    ~
    \begin{subfigure}[b]{0.23\textwidth}
        \includegraphics[width=\textwidth]{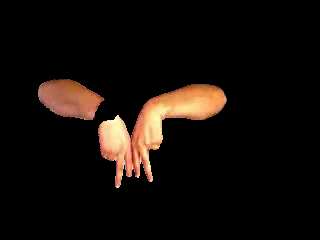}
        \caption{Hands segmentation}
        \label{fig:seg_example_hands_mask}
    \end{subfigure}
    ~
    \begin{subfigure}[b]{0.23\textwidth}
        \includegraphics[width=\textwidth]{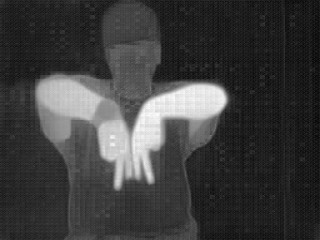}
        \caption{Hands logits\\ \ }
        \label{fig:seg_example_hands_logits}
    \end{subfigure}
    \caption{Examples of the body segmentation, body logits, hands segmentation, and hands logits as intermediate outputs of the segmentation streams in SegSLR.}
    \label{fig:streams_output}
\end{figure*}

Our new segmentation streams are integrated into the framework of~\cite{sarhan2020transfer} for ISLR classification. This results in SegSLR, as visible in Fig.~\ref{fig:system}. SegSLR consists of three main streams, one for the plain RGB data, one for optical flow data extracted from the RGB data using~\cite{zach2007duality}, and the aforementioned segmentation streams. As visible from Fig.~\ref{fig:system}, there exist four segmentation streams working on the outputs of the two promptable video segmentation modules for segmenting the signer's body and hands. The first stream uses masked RGB frames, effectively masking the background and focusing the processing on the signer based on the pose information. In contrast, the second stream processes the logits of the body segmentation. Similarly, for the two hands, the third and fourth streams do the same, processing masked RGB frames and the logits for both hands simultaneously. This focuses processing on the dominant parameter for ISLR, the hands. In case of the logits, the captured global context includes information about the location of the signer, making the relative location of the hands visible (see Fig.~\ref{fig:seg_example_hands_logits}). In contrast to most ISLR works focusing on hands, the precise and temporally consistent segmentations by SAM~2 capture the hand shapes, hand orientation, and even details about the fingers as visible in the examples in Fig.~\ref{fig:streams_output}. We evaluate the design choices regarding the streams in Sec.~\ref{sec:eval_streams}.

Finally, each segmentation stream in SegSLR uses an I3D CNN as commonly applied in ISLR literature~\cite{li2020transferring,sarhan2020transfer,hosain2021hand} to capture spatial and temporal dependencies. The per-stream results are combined using score-level fusion.


%

\subsection{Implementation Details}
SegSLR consists of three trainable components: (1) SAM~2, (2) RTMW, and (3) six I3D CNNs. For SAM~2, we apply a model pre-trained on SA-1B~\cite{kirillov2023segment} and SA-V~\cite{ravi2024sam} as proposed by~\cite{ravi2024sam} and do not add fine-tuning,  since segmentation annotations are unavailable in ISLR datasets. Similarly, we directly apply the pose estimation system RTMW, which is pre-trained on 14 datasets as suggested by~\cite{jiang2024rtmw}. For the I3D CNNs in SegSLR, we follow~\cite{sarhan2020transfer} and pre-train the I3D CNNs on ImageNet~\cite{russakovsky2015imagenet} and the Kinetics dataset~\cite{carreira2017quo}. Subsequently, each I3D CNN is fine-tuned individually with Adam optimizer, a batch size of 4, and early stopping with a patience of 3. We apply standard categorical cross-entropy loss per stream. When training or testing on videos, we uniformly sample 40 frames and extract a central $224\times 224$ crop from each frame. This ensures a constant size and length of the videos for the I3D CNNs. We additionally utilize data augmentation for training the I3D CNNs and shift the extracted crop horizontally or vertically and adjust the brightness~\cite{sarhan2020transfer}. 

\section{Evaluation}
We evaluate SegSLR on the commonly used ChaLearn249 IsoGD dataset~\cite{wan2016chalearn} comprising 249 gestures across 47,933 videos in complex environments and under challenging lighting conditions. For details about the covered gestures, we refer to~\cite{wan2016chalearn}.  We use the training dataset to train SegSLR and report results on the validation and test sets. Note that we train SegSLR five times on the training set and report the result with the median validation accuracy for a fair comparison. This model is also used to generate the test results. We additionally report the mean and standard deviation across the five trainings. We compare SegSLR against several recent ISLR methods utilizing RGB and/or pose information. This includes the baseline system of SegSLR~\cite{sarhan2020transfer}, methods that focus on the signer's body~\cite{sarhan2023pseudodepth,li2018gl} or hands~\cite{sarhan2023hands,li2018gl}, and a method combining RGB and pose information~\cite{li2018gl}. Note that we do not compare to methods utilizing depth, since this would be unfair, given the strong semantic cues of depth data for ISLR. A comparison to other methods like~\cite{gokcce2020score,hosain2021hand,gruber2021mutual} is impossible due to missing publicly available implementations. For assessing the quality, we use standard accuracy.


\subsection{Results on ChaLearn249 IsoGD Dataset}
Table~\ref{tab:results_chalearn} presents the results on the validation and test sets of ChaLearn249 IsoGD. The results clearly show that SegSLR outperforms all other methods on both sets by a substantial margin. This includes outperforming methods, which focus on the signer's hands through bounding boxes~\cite{sarhan2023hands,li2018gl} or the signer's body~\cite{sarhan2023pseudodepth,li2018gl}. The earlier explicitly shows the advantage of utilizing hand segmentations preserving hand shape details over simple boxes. Moreover, \cite{li2018gl}~also combine RGB and pose information by focusing their system on boxes around the hands, elbows, and shoulders of the signer. Comparing SegSLR to its baseline system, I3D-SLR, which only comprises the RGB stream and the optical flow stream, SegSLR shows an improvement of 9.21\% and 8.32\% in accuracy on validation and test sets. This is the result of adding the segmentation stream, combining RGB and pose information. Compared to the other variations of I3D-SLR, adding pseudo depth~\cite{sarhan2023pseudodepth}, focusing on moving areas~\cite{sarhan2021spatial}, and focusing on the signer's hands~\cite{sarhan2023hands}, SegSLR shows large improvements of up to 8.80\% and 6.56\% on the validation and test sets, respectively. Overall, this shows the strong performance of SegSLR based on combining pose and RGB information through promptable video segmentation.

\begin{table}[tb]
    \caption{Comparison of the proposed SegSLR to state-of-the-art ISLR methods on the ChaLearn249 IsoGD dataset. For SegSLR, we report the median accuracy on the validation set over five trainings, along with the respective mean and standard deviation in parentheses. $^\ast$: Numbers were not reported by the respective paper.}
    \begin{tabular}{lcc}
        \hline
        \multirow{2}{*}{\textbf{Method}} & \multicolumn{2}{c}{\textbf{\hspace*{9mm}Accuracy} (\%)} \\ & \textbf{Validation} & \textbf{Test} \\
        \hline
        C3D-LSTM~\cite{zhu2017multimodal} & 43.88 & -$^\ast$ \\
        SYSU ISEE~\cite{li2018gl} & 50.02 & -$^\ast$ \\
        XDETVP~\cite{zhang2017learning} & 51.31 & -$^\ast$ \\
        8-MFFs-3flc (5 crop)~\cite{kopuklu2018motion} & 57.40& -$^\ast$ \\
        I3D-SLR~\cite{sarhan2020transfer} & 62.09 & 64.44 \\
        I3D-pseudoDepth~\cite{sarhan2023pseudodepth} & 62.50 & 66.20 \\
        2SCVN-RGB-Fusion~\cite{duan2018unified} & 62.72& -$^\ast$ \\
        Hybrid Attn-I3D-SLR~\cite{sarhan2021spatial} & 65.02 & 68.89 \\
        TD-SLR~\cite{sarhan2023hands} & 67.13 & 70.91 \\
        \hline
        \multirow{2}{*}{SegSLR (ours)} & \textbf{71.30} & \textbf{72.76} \\
        & ($\mu=71.39$, $\sigma=0.41$) &  \\
        \hline
    \end{tabular}
    \centering
    \label{tab:results_chalearn}
\end{table}

\begin{figure*}[tb]
    \centering
    \begin{tabular}{cccccccc}
    
    \includegraphics[width=0.1\textwidth]{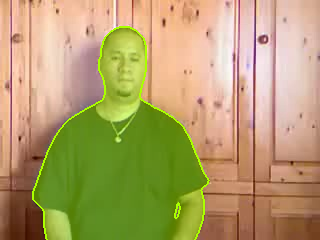} & \includegraphics[width=0.1\textwidth]{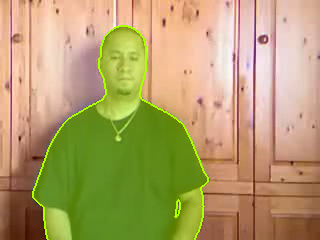} & \includegraphics[width=0.1\textwidth]{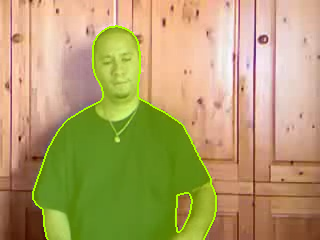} & \includegraphics[width=0.1\textwidth]{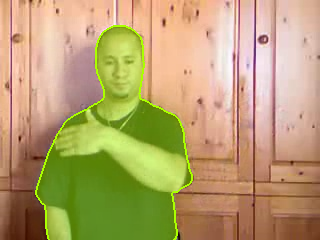} &
    \includegraphics[width=0.1\textwidth]{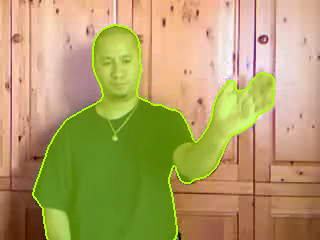} & \includegraphics[width=0.1\textwidth]{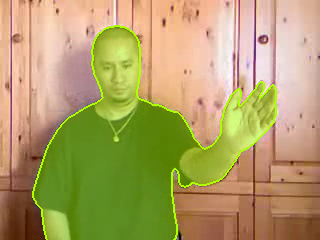}
    \includegraphics[width=0.1\textwidth]{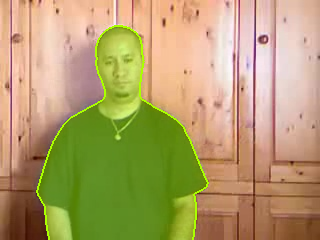} & \includegraphics[width=0.1\textwidth]{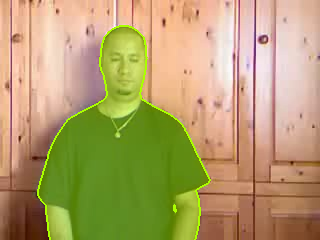}\\
    
    \vspace{2mm}
    \includegraphics[width=0.1\textwidth]{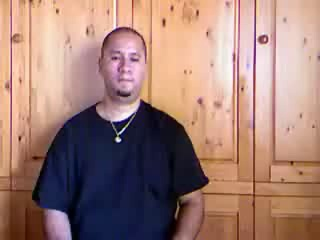} & \includegraphics[width=0.1\textwidth]{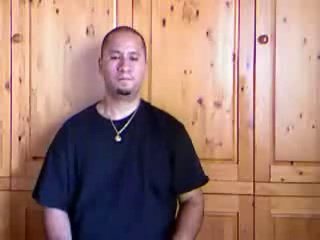} & \includegraphics[width=0.1\textwidth]{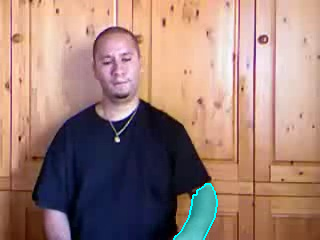} & \includegraphics[width=0.1\textwidth]{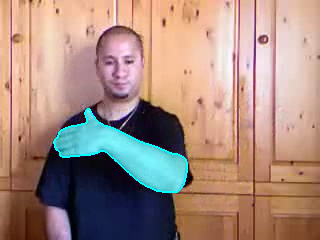} &
    \includegraphics[width=0.1\textwidth]{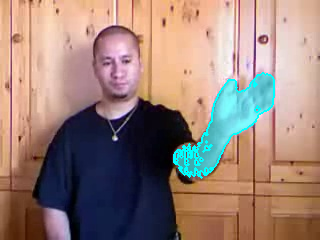} & \includegraphics[width=0.1\textwidth]{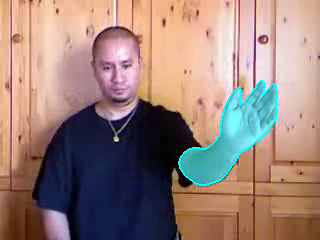}
    \includegraphics[width=0.1\textwidth]{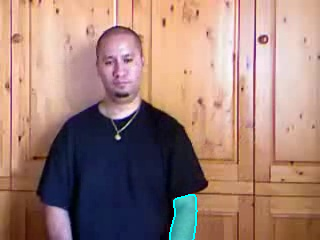} & \includegraphics[width=0.1\textwidth]{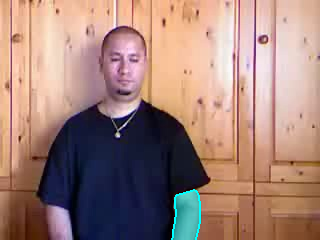}\\
    \includegraphics[width=0.1\textwidth]{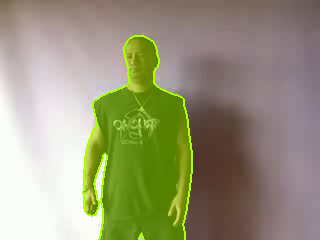} & \includegraphics[width=0.1\textwidth]{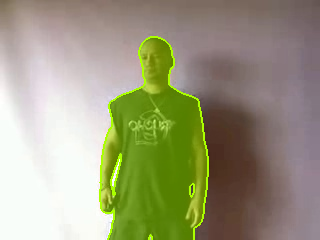} & \includegraphics[width=0.1\textwidth]{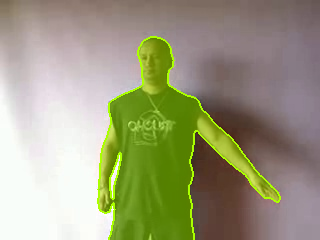} & \includegraphics[width=0.1\textwidth]{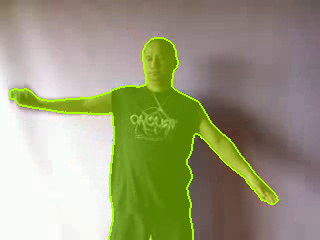} &
    \includegraphics[width=0.1\textwidth]{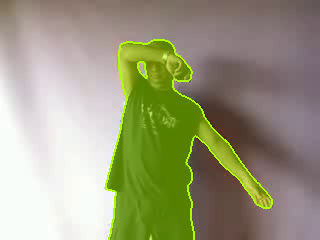} & \includegraphics[width=0.1\textwidth]{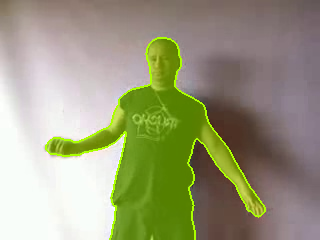}
    \includegraphics[width=0.1\textwidth]{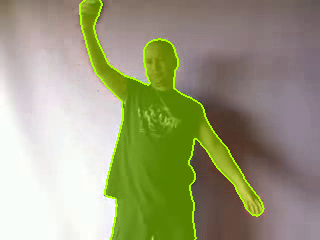} & \includegraphics[width=0.1\textwidth]{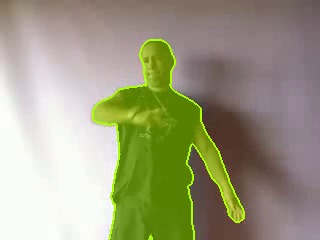}\\
    \vspace{2mm}
    \includegraphics[width=0.1\textwidth]{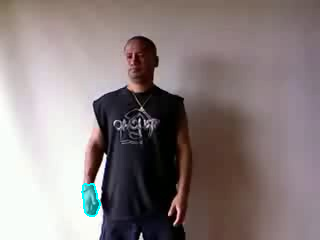} & \includegraphics[width=0.1\textwidth]{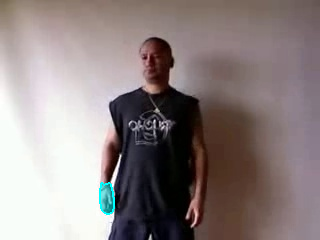} & \includegraphics[width=0.1\textwidth]{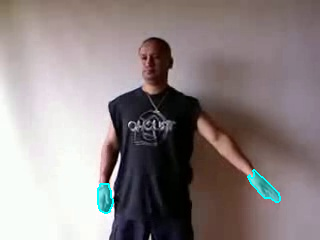} & \includegraphics[width=0.1\textwidth]{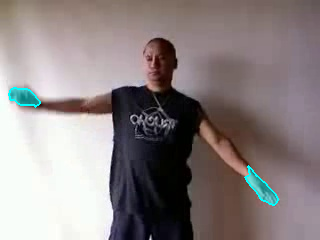} &
    \includegraphics[width=0.1\textwidth]{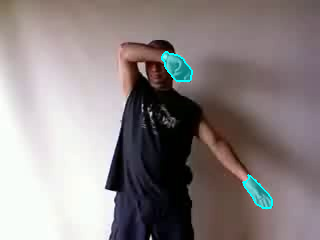} & \includegraphics[width=0.1\textwidth]{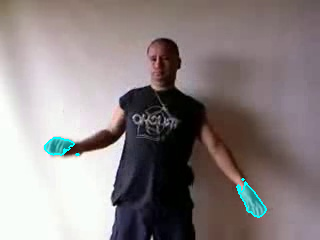}
    \includegraphics[width=0.1\textwidth]{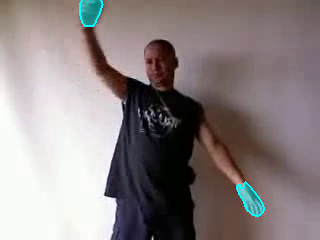} & \includegraphics[width=0.1\textwidth]{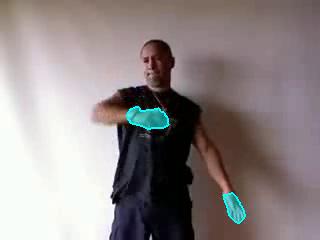}\\
    \includegraphics[width=0.1\textwidth]{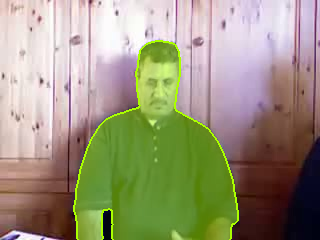} & \includegraphics[width=0.1\textwidth]{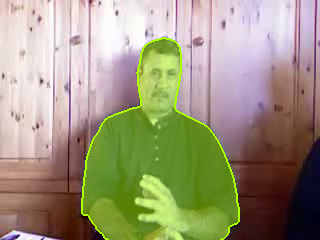} & \includegraphics[width=0.1\textwidth]{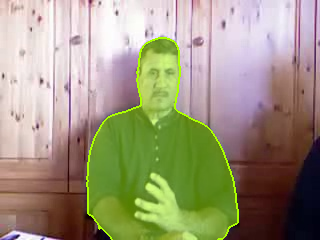} & \includegraphics[width=0.1\textwidth]{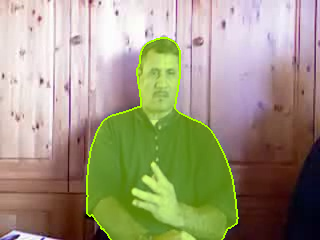} &
    \includegraphics[width=0.1\textwidth]{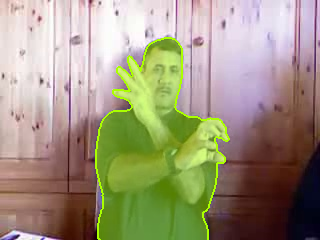} & \includegraphics[width=0.1\textwidth]{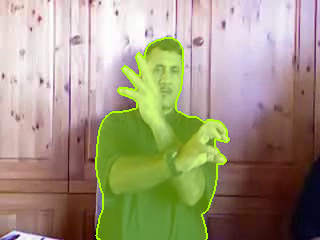}
    \includegraphics[width=0.1\textwidth]{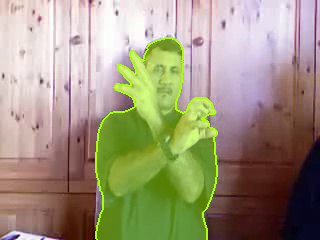} & \includegraphics[width=0.1\textwidth]{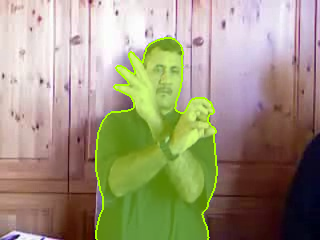}\\
    \includegraphics[width=0.1\textwidth]{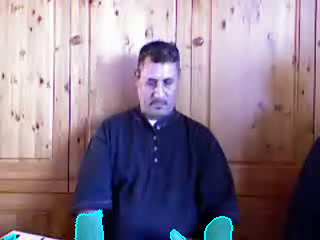} & \includegraphics[width=0.1\textwidth]{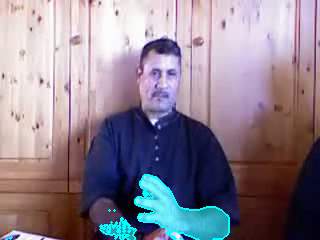} & \includegraphics[width=0.1\textwidth]{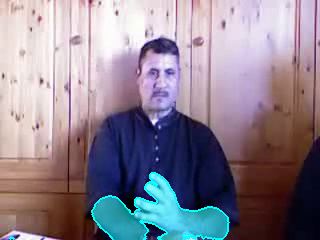} & \includegraphics[width=0.1\textwidth]{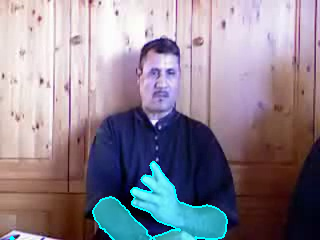} &
    \includegraphics[width=0.1\textwidth]{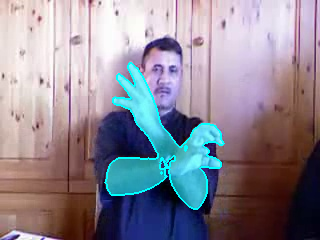} & \includegraphics[width=0.1\textwidth]{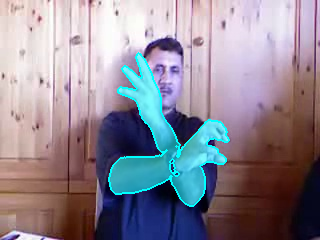}
    \includegraphics[width=0.1\textwidth]{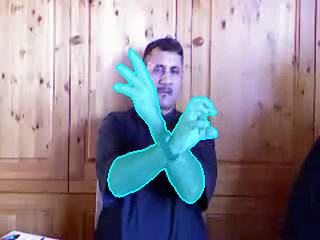} & \includegraphics[width=0.1\textwidth]{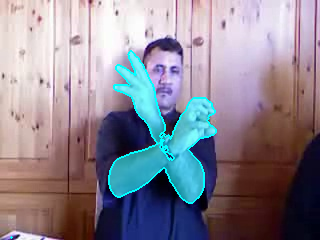}\\
    
    \end{tabular}
    \caption{Qualitative body~(green masks) and hands~(blue masks) segmentation results generated within our SegSLR on videos of the ChaLearn249 IsoGD test dataset. For each video, one sequence of frames is overlayed with the body segmentation and the hands segmentations, respectively. Note that we only show 8 of the 40 frames per video for brevity.}
    \label{fig:chalearn_segmentations}
\end{figure*}

The qualitative segmentation results in Fig.~\ref{fig:chalearn_segmentations} on ChaLearn249 IsoGD support the aforementioned quantitative results of SegSLR. Across all videos in Fig.~\ref{fig:chalearn_segmentations}, the segmentations of the signer's body and hands are accurate and suitable to focus the processing of SegSLR on these highly relevant areas. Even despite substantial hand movement or highly-textured backgrounds, the segmentations consistently capture the signer's body and hands. Specifically, in the videos in the first two rows, a waving hand is visible that challenges SAM~2 due to the high velocity of the movement. Yet, the segmentations are consistent. In the second and third rows, the videos show glosses where a hand is moved in front of the face. Despite visual similarity between the hands and the face, SAM~2, prompted with our point prompts derived from pose information, distinguishes between hands and face. This is due to the negative prompts covering the face when generating the hand segmentation. Finally, the last row also shows an example of a sign with a complex finger posture. Still, the hand segmentation in SegSLR captures all fingers in detail, given the pose-based point prompts. 

%

\begin{table}[tb]
    \centering
    \caption{Results of SegSLR on the ChaLearn249 IsoGD dataset with three versions for the segmentation streams. Body\textsubscript{\textit{RGB}} denotes the segmentation stream utilizing masked RGB frames. Body\textsubscript{\textit{Logits}} is the segmentation stream processing the raw logits of the body segmentation. Hands\textsubscript{\textit{RGB}} and Hands\textsubscript{\textit{Logits}} are the respective streams segmenting the hands. }
    \begin{tabular}{lcc}
        \hline
        \multirow{2}{*}{\textbf{Input Streams}} & \multicolumn{2}{c}{\textbf{Accuracy} (\%)} \\ & \textbf{Validation} &  \textbf{Test} \\
        \hline
        Base & 62.09 & 64.42 \\
        + Body\textsubscript{\textit{RGB}} & 65.51 & 67.33 \\
        + Body\textsubscript{\textit{RGB}} + Body\textsubscript{\textit{Logits}} & 67.10 & 69.78 \\
        + Body\textsubscript{\textit{RGB}} + Body\textsubscript{\textit{Logits}} + Hands\textsubscript{\textit{RGB}} + Hands\textsubscript{\textit{Logits}} & \textbf{71.30} & \textbf{72.76 }\\
        \hline
    \end{tabular}
    \label{tab:input_streams}
\end{table}

\subsection{Ablation Studies}


\subsubsection{Input Streams}
\label{sec:eval_streams}
To assess the impact of SegSLR's focus on the signer's body and hands, and the importance of utilizing both the binary segmentation mask and the logits, we present in Tab.~\ref{tab:input_streams} the step-by-step results, from the baseline system~\cite{sarhan2020transfer} to SegSLR. In each step, we add segmentation streams to combine pose and RGB information. Given the baseline, we first add SegSLR's body segmentation by masking the respective RGB frame~(Body\textsubscript{\textit{RGB}} in Tab.~\ref{tab:input_streams}). The improvements of 3.42\% and 2.91\% clearly indicate the advantage of an additional focus on the signer's body. Adding the logits of the body segmentation in another stream~(Body\textsubscript{\textit{Logits}} in Tab.~\ref{tab:input_streams}) further improves the results by 1.59\%/2.45\% and shows the additional value of the logits. Finally, adding the same segmentation streams for the hands (Hands\textsubscript{\textit{RGB}} and Hands\textsubscript{\textit{Logits}} in~Tab.~\ref{tab:input_streams}), further improves the results by 4.22\% and 2.98\%, highlighting the importance of hands for ISLR. Overall, the results in Tab.~\ref{tab:input_streams} clearly show the value of all segmentation streams combining RGB and pose information.

%
%

\subsubsection{Segmentation Method}
\begin{figure}[tb]
    \centering
    \begin{tabular}{cccc}
    \includegraphics[width=0.22\textwidth]{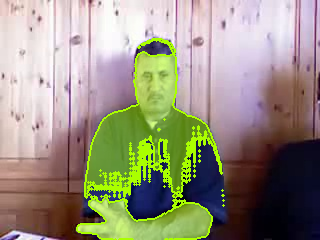} & \includegraphics[width=0.22\textwidth]{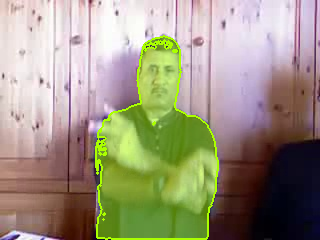} & \includegraphics[width=0.22\textwidth]{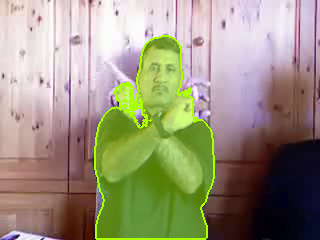} & \includegraphics[width=0.22\textwidth]{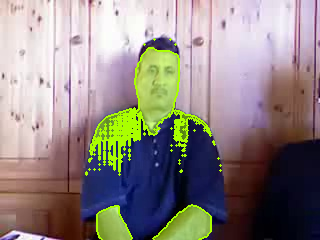}  \\
    \includegraphics[width=0.22\textwidth]{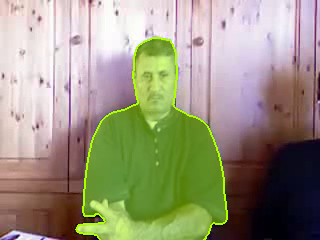} & \includegraphics[width=0.22\textwidth]{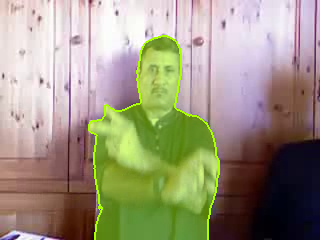} & \includegraphics[width=0.22\textwidth]{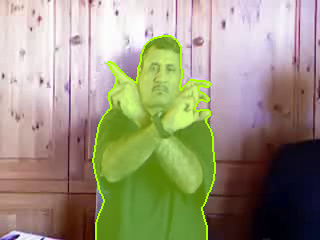} & \includegraphics[width=0.22\textwidth]{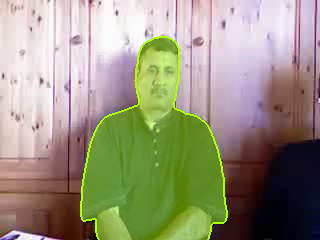}  \\
    \end{tabular}
    \caption{Comparison of segmentations of the signer's body with SAM~(upper row) and SAM~2~(lower row) in SegSLR on four frames of a video from the ChaLearn249 IsoGD test dataset.}
    \label{fig:sam1_vs_sam2}
\end{figure}

We present the results of SegSLR with three different segmentation methods: well-known Mask R-CNN~\cite{he2018mask} trained on the COCO dataset~\cite{lin2014microsoft} to segment only humans, denoted as Mask R-CNN\textsubscript{Person}, SAM~\cite{kirillov2023segment}, and SAM~2~\cite{ravi2024sam} as in the proposed SegSLR. Note that we utilize only the segmentation stream for the signer's body in SegSLR to match Mask R-CNN's training on COCO. For SAM and SAM~2, we apply the same keypoints as described in Sec.~\ref{sec:prompt} The results in Tab.~\ref{tab:body_comparison} show that SegSLR outperforms the baseline with any segmentation stream. Yet, SAM and SAM~2 surpass Mask R-CNN by up to 3.20\%, which underlines the strong zero-shot segmentation ability of these foundation models. Comparing SAM and SAM~2 shows that SAM~2 leads to an improvement of 1.37\% and 0.87\%. A major reason for this is the temporal consistency of the segmentations between the frames. This is also visible in the qualitative results in Fig.~\ref{fig:sam1_vs_sam2}, where the SAM segmentations~(upper row) flicker substantially between the frames, while the SAM~2 segmentations~(lower row) consistently cover the entire body.


\begin{table}[tb]
    \caption{Results of SegSLR with different segmentation methods on the ChaLearn249 IsoGD dataset. Note that SegSLR only includes the body segmentation stream~(Body\textsubscript{\textit{RGB}}) here.}
    \begin{tabular}{lcc}
        \hline
        \multirow{2}{*}{\textbf{Input Streams}} & \multicolumn{2}{c}{\textbf{Accuracy} (\%)} \\ & \textbf{Validation} & \textbf{Test}\\
        \hline
        Base & 62.09 & 64.42 \\
        Base + Mask R-CNN\textsubscript{Person} & 62.31 & 65.44 \\
        Base + SAM (Body\textsubscript{\textit{RGB}} only) & 64.14 & 66.46 \\
        Base + SAM~2 (Body\textsubscript{\textit{RGB}}  only) & 65.51 & 67.33\\
        \hline
    \end{tabular}
    \centering
    \label{tab:body_comparison}
\end{table}

\section{Conclusion}
Effectively utilizing both RGB and pose information is key for high-quality ISLR. To address this combination of RGB and pose information without losing details important to understand sign language, we proposed the novel ISLR system SegSLR. It innovatively combines RGB and pose information through promptable video segmentation using pose keypoints to prompt SAM~2 and detect the signer's body and hands for a focused processing of the RGB data. The strength of this novel design for ISLR is supported by our experiments on the ChaLearn249 IsoGD dataset, where SegSLR outperforms all competing methods. Our ablation studies also validated the focus of SegSLR on the signer's body and hands as well as the use of SAM~2 based on prompts from pose estimation data. Overall, SegSLR presents another step to bridge the communication gap between people inside and outside the deaf or hard-of-hearing community.

\bibliographystyle{splncs04}
\bibliography{021-main.bib}

\end{document}